\documentclass[
]{ceurart}

\usepackage{listings}
\usepackage{amsmath,amssymb,amsfonts}
\usepackage{algorithmic}
\usepackage{graphicx}
\usepackage{textcomp}
\usepackage{xcolor}
\usepackage{url}
\usepackage{array}
\usepackage{multirow}
\usepackage{multicol}
\usepackage{caption} 

\newcommand{\puck}{Pucktrick\xspace}
\newcommand{\ml}{machine learning\xspace}
\newcommand{\sdg}{synthetic data generation\xspace}
\begin{document}
\copyrightyear{2025}
\copyrightclause{Copyright for this paper by its authors.
  Use permitted under Creative Commons License Attribution 4.0
  International (CC BY 4.0).}

\conference{SEBD 2025: 33st Symposium on Advanced Database Systems.  June 16-19 Ischia, Campania,  Italy }

\title{PuckTrick: A Library for Making Synthetic Data More Realistic}


\author[1]{Alessandra Agostini}[%
email=alessandra.agostini@unimib.it,
]
\author[1]{Andrea Maurino}[%
email=andrea.maurino@unimib.it,
]
\author[1]{Blerina Spahiu}[%
email=Blerina.Spahiu@unimib.it,
]

\address[1]{Department of Informatics, Systems and Communication\\ University of Milan -Bicocca\\
Viale Sarca 336, 20126 Milan, Italy \\
}

\maketitle

\begin{abstract}
The increasing reliance on machine learning (ML) models for decision-making requires high-quality training data. However, access to real-world datasets is often restricted due to privacy concerns, proprietary restrictions, and incomplete data availability. As a result, synthetic data generation (SDG) has emerged as a viable alternative, enabling the creation of artificial datasets that preserve the statistical properties of real data while ensuring privacy compliance. Despite its advantages, synthetic data is often overly clean and lacks real-world imperfections, such as missing values, noise, outliers, and misclassified labels, which can significantly impact model generalization and robustness. To address this limitation, we introduce \puck, a Python library designed to systematically contaminate synthetic datasets by introducing controlled errors. The library supports multiple error types, including missing data, noisy values, outliers, label misclassification, duplication, and class imbalance, offering a structured approach to evaluating ML model resilience under real-world data imperfections. \puck provides two contamination modes: one for injecting errors into clean datasets and another for further corrupting already contaminated datasets. Through extensive experiments on real-world financial datasets, we evaluate the impact of systematic data contamination on model performance. Our findings demonstrate that ML models trained on contaminated synthetic data outperform those trained on purely synthetic, error-free data, particularly for tree-based and linear models such as SVMs and Extra Trees.
\end{abstract}


\section{Introduction}
\label{ref:intro}

The rapid advancement of technology in recent years has facilitated the efficient and large-scale gathering, processing, and analysis of data for informed decision-making across various domains. As of 2024, approximately 402.74 million terabytes (or 0.4 zettabytes) of data are created each day, totaling around 147 zettabytes annually\footnote{\url{https://explodingtopics.com/blog/data-generated-per-day}}. This daily data generation is projected to increase to about 181 zettabytes per year by 2025\footnote{\url{https://www.statista.com/statistics/871513/worldwide-data-created/}}. The rapid evolution of statistical analysis and pattern recognition techniques has revolutionized the ability to extract meaningful insights from such data. However, the effectiveness of these methods depends on the quality of the data being analyzed. When data is incomplete, inconsistent, duplicated, or lacks proper security measures, the accuracy and reliability of the results are significantly compromised. 

One of the major challenges in \ml is the availability of real dataset for training the model \cite{paleyes2022challenges}. In many applied contexts, it is not always possible to use real data due to privacy concerns or the unwillingness of data owners to share their proprietary information. To address both data quality and privacy concerns, synthetic data has emerged as a trade-off solution \cite{figueira2022survey}. Synthetic data replicates the statistical properties of real data while eliminating direct exposure of sensitive information. It also helps mitigate the issue of missing values by generating artificial replacements, balancing datasets by creating synthetic examples for underrepresented groups, and providing an added layer of security since the generated data does not correspond to real-world individuals or entities. Existing models focus on creating synthetic data by leveraging techniques such as generative adversarial networks (GANs)\cite{9489160}, variational autoencoders (VAEs)\cite{9171997}, and statistical sampling methods \cite{liu2022real} to replicate the properties of real-world data distributions.

Once a synthetic dataset is generated, its quality is typically assessed by training a preliminary \ml model and evaluating performance metrics such as accuracy, F1-score, precision, and recall. These metrics help determine whether the synthetic data preserves meaningful patterns from real-world data. By performing this initial assessment with a lightweight model or proxy task, researchers can estimate the dataset's utility before committing to the expensive and time-consuming process of full-scale model training \cite{iskander2024quality}. However, this approach overlooks a critical aspect: real-world data is inherently noisy and imperfect. For example, a medical sensor may malfunction and provide incorrect measurements. A change in the IT system that feeds the dataset could result in missing data or values on a different scale than the original ones. Finally, experts might misclassify certain data points. Synthetic datasets, typically designed to be clean and well-structured, may fail to capture these real-world inconsistencies. As a result, models trained on idealized synthetic data risk poor generalization when applied to real-world scenarios, where unpredictable data variations and errors are common \cite{chen2024unveiling}. This gap highlights the importance of incorporating realistic noise and imperfections into synthetic data to improve its applicability in practical machine-learning applications.

To address this limitation, we developed a python library called \puck designed to systematically and controllably contaminate datasets, thereby generating more realistic training data. The \puck library simulates errors that can be encountered in a dataset used for training models. The most common types of errors encountered in real-world datasets for \ml training include:

\begin{itemize}
    \item \textit{Missing Data}: Instances where values are absent, leading to biases and potential inaccuracies in predictions.
    \item \textit{Noisy Data}: Random errors or fluctuations in the data that obscure underlying patterns and degrade model performance.
    \item \textit{Outliers:} Data points that significantly deviate from the norm, potentially skewing model predictions.
    \item \textit{Imbalanced Data}: Unequal representation of classes, leading to bias in favor of the majority class.
    \item \textit{Duplicate Data}: Repeated or nearly identical entries that can lead to overfitting and inefficient training.
    \item \textit{Incorrect Labels}: Mislabelled instances that negatively impact classification performance and model reliability.
\end{itemize}

The \puck library offers two modes of operation. The first mode introduces errors into a dataset that is initially considered clean, allowing users to simulate real-world imperfections systematically. The second mode is designed for cases where a dataset is already contaminated but requires further corruption with specific types of errors. In such situations, the extended methods of \puck can be used to incrementally introduce additional errors until the desired error percentage is reached. Beyond dataset contamination, \puck also serves other purposes. For instance, researchers developing data-cleaning techniques can leverage \puck to inject specific errors into a dataset and then assess whether their method effectively detects and corrects these errors. This makes \puck a versatile tool for both generating realistic training data and benchmarking data-cleaning algorithms.


Experiments show that training \ml models on synthetic data with controlled errors (introduced using \puck) results in better accuracy compared to training on purely synthetic, error-free data. This confirms that exposing models to realistic imperfections during training makes them more robust and adaptable, ultimately improving their performance in real-world scenarios.

Given the importance of training \ml models with realistic, imperfect data, in this study, we investigate the impact of controlled dataset contamination on \ml model performance by analyzing:

\begin{itemize}
    \item The effect of training on contaminated synthetic data versus purely synthetic, error-free data,
    \item The influence of different types of data errors on model generalization and real-world applicability,
    \item The extent to which systematic error introduction enhances a model’s ability to handle naturally occurring data imperfections, and
    \item The potential of \puck as a benchmarking tool for evaluating data-cleaning techniques and error-handling strategies in \ml workflows.
\end{itemize}

The paper is organized as follows: section \ref{ref:related} reviews the state of the art in both \sdg and data contamination. Section \ref{ref:archi} introduces the \puck library and its functionalities while section \ref{ref:exp} outlines the experimental setup and discusses the preliminary results.
Finally, section \ref{ref:conclusion} provides the conclusions and explores potential directions for future work.

\section{Related Work}
This section explores the current state of the art in SDG techniques, approaches for introducing controlled data imperfections, and their implications for model robustness and real-world applicability.

\label{ref:related}
\subsection{Synthetic data generation algorithms}
Synthetic data generation (SDG) involves creating artificial datasets that replicate the statistical properties of real-world data. SDG is relevant in various scenarios where real data cannot be used for training machine learning models. In the medical domain, for instance, utilizing real patient data is challenging due to confidentiality concerns. Similarly, in the financial sector, datasets for fraud detection are often highly imbalanced, with a predominance of non-fraudulent transactions.
Synthetic data is obtained from a generative process based on properties of real data. A comprehensive survey \cite{bauer2024comprehensive} analyzed 417 SDG models developed over the past decade, highlighting the evolution of model performance and complexity. The study found that neural network-based approaches, particularly Generative Adversarial Networks (GANs), dominate the landscape, especially in computer vision applications. Emerging models like diffusion models and transformers are also gaining traction, offering promising avenues for future research.

The most common format for synthetic data generation is the tabular structure, which consists of columns (also referred to as features) and rows (also called observations).  According to \cite{Fonseca2023} a synthetic data generation process can be described along four different dimensions: \textit{Architecture} (it represents the type of data augmentation technique used), \textit{Application level} (it refers to the phase of \ml pipeline where the process is included), \textit{Scope} (it is related to the usage of existing dataset properties), and \textit{Data Space} (it is related to the representation model used in the process).
The primary metric for assessing the quality of synthetic data is its ability to enhance \ml model performance, typically measured through accuracy or F1-score. Evaluating synthetic data quality before full-scale model training is crucial, as training an ML classifier can be computationally expensive and time-consuming. By anticipating data quality early, researchers can optimize resources and improve model efficiency.


Generative Adversarial Networks (GANs) have been extensively explored for generating synthetic tabular data. These models consist of a generator and a discriminator that engage in a minimax game, leading to the production of data that closely resembles real datasets. Recent studies have adapted GANs to handle the unique challenges of tabular data, such as mixed data types and complex feature dependencies. Among others frequently used \sdg algorithm are CTGAN \cite{Xu2019}, GReaT \cite{DBLP:conf/iclr/BorisovSLPK23}, SDV \cite{SDV}.
Conditional Tabular GAN (CTGAN) is a deep generative model specifically designed for the synthesis of tabular data. Unlike traditional GANs, which struggle with the complex statistical properties of tabular datasets, CTGAN introduces several innovations to enhance data fidelity. To address the issue of managing at the same time continuous and categorical features, CTGAN employs a mode specific normalization technique based on a Variational Gaussian Mixture Model (VGM) \cite{1658297}, which effectively encodes continuous variables while preserving their original distribution.  CTGAN also mitigates the underrepresented class problem by incorporating a conditional generator, ensuring that all categories are sufficiently represented during training. Furthermore, the model employs a training-by-sampling approach, where data instances are selected based on the log-frequency of categorical values rather than uniformly. This strategy improves the generator’s ability to produce balanced and representative synthetic samples. The network architecture of CTGAN consists of fully connected neural networks for both the generator and the critic. To stabilize training and improve the quality of the generated data, the model utilizes the Wasserstein GAN with Gradient Penalty (WGAN-GP) framework \cite{Walia2020SynthesisingTD, pmlr-v178-milne22a}. The generator takes as input a random noise vector along with a conditional variable and produces synthetic tabular records. The critic then evaluates these records, helping the model refine its ability to generate realistic samples.

Variational Autoencoders (VAEs) have also been applied to tabular data synthesis, focusing on learning latent representations that capture the underlying data distribution. By sampling from the latent space, VAEs can generate new data points that maintain the statistical properties of the original dataset. Additionally, models like TVAE (Tabular Variational Autoencoder) \cite{Fonseca2023} have been developed to specifically address the challenges of tabular data, incorporating mechanisms to handle diverse data types and complex relationships between features.

The role of data-centric AI in improving synthetic tabular data generation and evaluation is investigated in \cite{hansen2023reimagining}. Traditional approaches rely on statistical fidelity metrics to assess synthetic data quality, but the authors argue that these methods alone are insufficient. Instead, they propose incorporating data profiling techniques that categorize samples based on learning difficulty to guide synthetic data generation. Through extensive benchmarking across eleven datasets and five state-of-the-art synthetic data generators, the study demonstrates that considering data profiles enhances model performance, model selection reliability, and feature selection effectiveness. The findings suggest that different generative models excel in different tasks, emphasizing the need for task-specific synthetic data evaluation.

Emerging approaches have explored the use of language models for generating synthetic tabular data. By treating rows as sequences, these models can capture dependencies between features effectively.
The GReaT model \cite{DBLP:conf/iclr/BorisovSLPK23} utilizes large language models (LLMs) for synthetic tabular data generation. Instead of encoding tabular data numerically, GReaT transforms each row into a structured textual representation using a subject-predicate-object format, maintaining semantic coherence. A pretrained transformer-based LLM is then fine-tuned on this transformed dataset, enabling it to generate synthetic tabular samples while preserving statistical properties. To ensure flexibility, a random feature order permutation step is introduced, preventing the model from learning an artificial ordering among features. The model supports arbitrary conditioning, allowing data generation based on selected feature constraints. During inference, GReaT generates synthetic records by iteratively sampling feature values in an autoregressive manner. The generated data is then transformed back into tabular form through pattern-matching algorithms.  A python library\footnote{\url{https://github.com/kathrinse/be_great}} is also available. 

The development of automated platforms for synthetic data generation has streamlined the process of creating high-quality synthetic datasets. For example, the Synthetic Tabular Neural Generator (STNG) \cite{rashidi2024novel} integrates multiple generation methods with an AutoML module, facilitating the automatic generation of synthetic data tailored to specific tasks. This platform addresses the need for user-friendly tools that can adapt to various data characteristics and requirements. 

The Synthetic Data Vault (SDV) \cite{SDV} is a \sdg model designed to automatically generate synthetic data, enabling data science applications while preserving privacy. SDV builds generative models of relational databases, allowing the synthesis of artificial data that retains the statistical and structural properties of real datasets. The system employs a recursive modeling technique called Recursive Conditional Parameter Aggregation (RCPA), which models the relationships between database tables to generate realistic synthetic data. SDV has been evaluated on multiple publicly available datasets, demonstrating that synthetic data can effectively replace real data in predictive modeling tasks. 
Its generative process incorporates multivariate statistical techniques, including Gaussian Copulas, to model data distributions accurately. SDV is now implemented in a python library called SDV\footnote{\url{https://github.com/sdv-dev/SDV}}  and it includes other \sdg models such as CTGAN, and Copulas, allowing users to choose the most appropriate approach for their specific data characteristics.

Assessing the quality and utility of synthetic tabular data is crucial for its adoption in practical applications \cite{liu2024scaling}. Structured evaluation frameworks have been proposed to provide a comprehensive assessment of synthetic data generators. These frameworks consider various metrics, including resemblance, utility, and privacy preservation, offering a standardized approach to evaluate and compare different synthetic data generation methods. 

\subsection{Data contamination}
To the best of the authors' knowledge, no study has attempted to define a systematic approach for introducing errors into the dataset to realistically simulate data. In some papers, errors are introduced to test the proposed solution only. 
For example, in \cite{SDV} authors introduced two types of noise: table and key noise. The table noise method alters the covariance structure between variables in the dataset. To introduce noise, the SDV modifies the covariance values $\sigma_{ij}$ (where $i \neq j$) by halving them, effectively reducing the strength of correlations between features. This perturbation ensures that the synthetic data maintains its general statistical properties but with weaker dependencies, making it more distinct from the original dataset. The key noise method affects the integrity of foreign key relationships in relational datasets. Instead of using the inferred relationships when synthesizing child tables, the SDV randomly assigns foreign keys to synthetic records. This disrupts the original structural dependencies within the dataset, introducing additional randomness while preserving the overall schema. Such methods are only used to test the quality of the proposed \sdg algorithms and there are no available noise functions in the python library. 

Numerous studies have investigated the impact of the most common types of data errors on machine learning models, and an equally large number of approaches have been proposed to detect and correct these errors. In \cite{NIPS2013_3871bd64}, authors address the problem of binary classification in the presence of random label noise, where training labels may be independently flipped with a certain probability. The authors propose two approaches to adapt surrogate loss functions for robust learning. These methods lead to a key result: widely used techniques such as weighted SVMs and weighted logistic regression are provably noise-tolerant. Authors of \cite{Emmanuel2021} discuss a survey of existing methods for managing the missing data problem in \ml. In \cite{Boukerche2020OutlierDetection} a survey of outlier detection techniques reports seven  different classes of outlier detection techniques for a total of 47 different proposed techniques.

\section{The \puck Library}
\label{ref:archi}
To systematically introduce artificial errors into datasets and simulate real-world data imperfections, we developed Pucktrick\footnote{Puck is the name of the elf in “A Midsummer Night’s Dream” by William Shakespeare, who is famous for enjoying causing trouble and playing tricks on mortals and other fairies alike}\footnote{\url{https://andreamaurino.github.io/pucktrick-ui-docs/}}. This tool enables users to contaminate datasets with controlled levels of errors as the ones described in Section \ref{ref:intro}, ensuring a structured and reproducible approach to studying the effects of data corruption. 


The Pucktrick library is designed to introduce errors at a specified percentage, offering two operational modes:
\begin{itemize}
    \item \textit{New mode}: This mode introduces errors into a clean dataset, allowing users to inject a predefined level of corruption.
    \item \textit{Extended mode}: This mode further contaminates a dataset that has already been modified, ensuring that additional noise is selectively introduced into previously untouched portions while maintaining the desired overall error distribution.
\end{itemize}
The library consists of five distinct modules, each designed to introduce a specific category of data errors. These errors can be applied either at the dataset level (affecting all records) or at the feature level (targeting specific attributes). The supported data types include:

\begin{itemize}  
    \item \textit{Categorical data}: Errors such as label flipping, misspellings, and misclassifications.
    \item \textit{Numerical data}: Noise injection, incorrect scaling, and value swapping.
    \item \textit{Boolean data}: Random inversions of true/false values.
    \item \textit{Date and time data}: Temporal shifts, incorrect formatting, and missing timestamps.
\end{itemize}

By allowing users to precisely control the level and type of errors introduced, Pucktrick serves as a powerful tool for evaluating machine learning robustness, benchmarking data-cleaning algorithms, and understanding the impact of real-world noise on analytical models. The library provides a structured, repeatable, and scalable approach to data contamination, bridging the gap between synthetic data generation and real-world dataset imperfections.

In the following subsection, we introduce the \puck modules. Readers seeking more details can refer to the online documentation for further information.

\subsection{Duplicate data}
The \textit{duplicate} module offers functionalities for duplicating rows within a dataset, available in both 'new' and 'extended' modes. This can be done either randomly across the entire dataset or in a targeted manner, where specific rows are duplicated based on a chosen attribute value. This feature is particularly useful for handling imbalanced datasets, where oversampling a particular class can improve the performance of multi-class classifiers. This module allows users to test how well machine learning models handle redundant information and potential data leakage scenarios.

\subsection{Label misclassification}
The \textit{label} module introduces errors in target variables, simulating real-world misclassification errors that commonly occur in practical settings \cite{frenay2014comprehensive}. 
In binary classification tasks, the module allows direct label swapping (e.g., converting 0 to 1 and vice versa). For multi-class classification, labels are randomly reassigned to a different class within the dataset. This feature enables researchers to study the impact of label noise on classification performance and evaluate the effectiveness of error correction algorithms.

\subsection{Missing data}
The \textit{missing} module includes methods for inserting "null" values in a specific column with a predefined percentage. Missing values are a frequent challenge in real-world datasets, often caused by sensor failures, human errors, or incomplete data collection. This module enables controlled experimentation with various missing data scenarios, helping to assess the performance of imputation techniques and machine learning models under different levels of data incompleteness.

\subsection{Noisy data}
The \textit{noise} is one of the most complex components of the library, allowing users to introduce random distortions into dataset features. Different methods are implemented depending on the data type: For each data type outlined in Section \ref{ref:archi}, dedicated methods have been implemented.

\begin{itemize}
    \item Continuous and discrete features: Noise is generated from a normal distribution within the range defined by the minimum and maximum values of the feature. This simulates real-world measurement errors or rounding imprecisions.
    \item Categorical integer features: Original values are replaced with alternative values randomly sampled from the existing set, following a normal distribution to ensure realistic variation.
    \item Categorical string features: The module generates synthetic categories by introducing new, normally distributed string values. This simulates scenarios where new categories emerge in real-world data (e.g., evolving customer preferences, new product categories).
\end{itemize}


\subsection{Outlier data}
The \textit{outlier} module generates extreme data points to simulate anomalies and rare events in a dataset. The approach differs based on the data type:

\begin{itemize}
    \item Continuous features: The module employs the 3-sigma method, a statistical technique for detecting and generating outliers under the assumption of a normal distribution. The method calculates the mean (\(\bar{X}\)) and standard deviation (\(\sigma\)) of the feature define the outlier boundaries as: \[\text{Upper Boundary} = \bar{X} + 3\sigma, \quad \text{Lower Boundary} = \bar{X} - 3\sigma\]
    \item Integer features: A similar 3-sigma methodology is applied to discrete numerical data.
    \item Categorical features: To introduce categorical outliers, the module simulates an unknown category by adding values that do not exist in the dataset. These values can take the form of:
    \begin{itemize}
        \item A new string category labeled "Puck was here", mimicking unexpected input values.
        \item An integer category not previously present in the dataset, representing unseen class labels in classification tasks.
    \end{itemize}
\end{itemize}

\section{\puck evaluation}
\label{ref:exp}

\subsection{Evaluation pipeline}
\begin{figure}[h]
    \centering
    \includegraphics[width=0.65\linewidth]{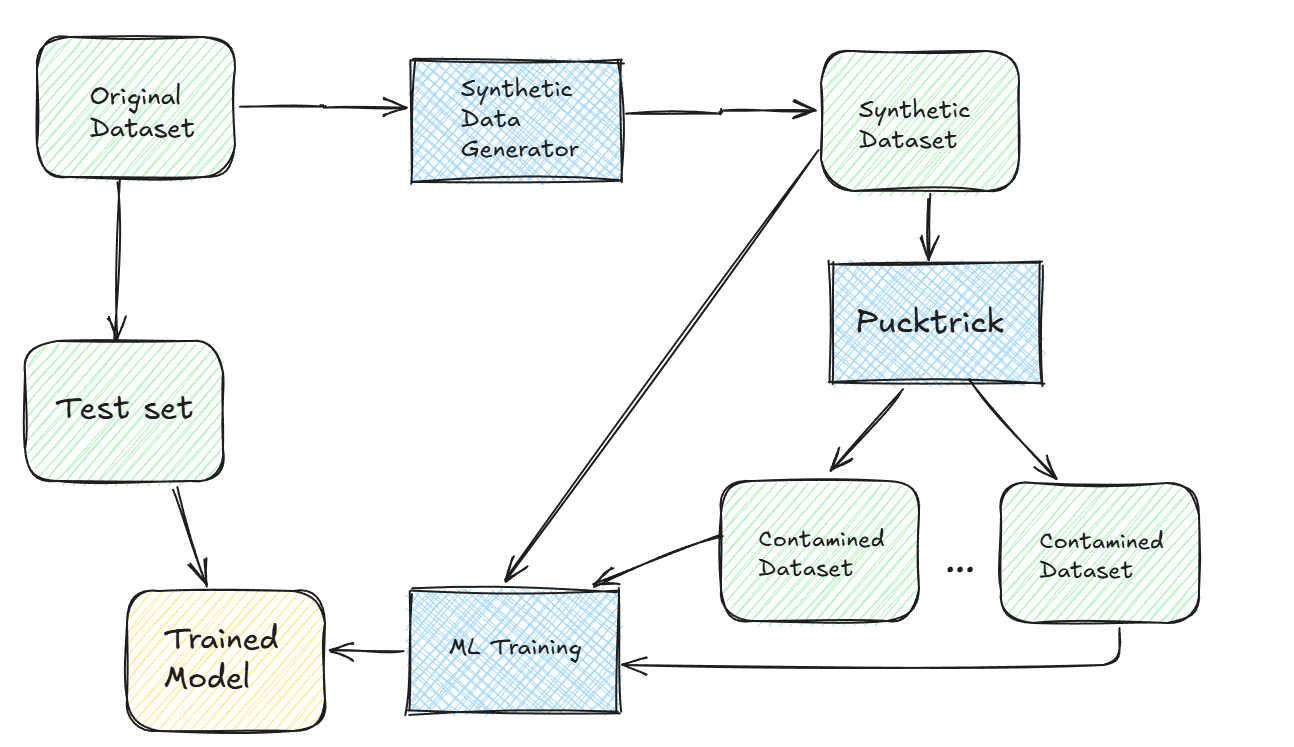}
    \caption{The proposed pipeline}
    \label{fig:archi}
\end{figure}

To assess the effectiveness of the \puck library, we construct a dedicated pipeline that highlights the advantages of introducing errors into synthetically generated data. Figure \ref{fig:archi} provides an overview of the pipeline. The provided pipeline illustrates the process of generating, contaminating, and utilizing synthetic data for \ml training. 
It begins with an original dataset, which is split into two parts: one used for generating synthetic data and another reserved as a test set. The synthetic data generator produces a synthetic dataset by applying one of the \sdg algorithms referenced in Section \ref{ref:related} to produce a synthetic dataset that retains the statistical properties of the original data.  
Subsequently, this synthetic dataset is passed through  the \puck library to introduce controlled noise and data contamination as outlined in Section \ref{ref:archi}. 

To highlight the contribution of each type of error produced, the pipeline constructs a separate contaminated dataset for each introduced error type (e.g., labels misclassification, outliers, etc.). The contaminated datasets, along with the synthetic dataset, are used to train one or more \ml models. Once training is completed, the resulting models are then employed to predict the correct class using the test set of the original dataset, which serves as a benchmark to compare the model’s performance under different training conditions. The obtained results are subsequently compared with the available labels in the test set to compute standard performance metrics, such as accuracy, F1-score, precision, recall, and AUC. Finally, the different performance outcomes are compared to assess the effectiveness of the proposed approach. This approach ensures a systematic evaluation of data contamination effects on model generalization, providing insights into how well \ml models handle real-world noisy and imperfect data.

\begin{figure}[h]
    \centering
    \includegraphics[width=0.7\linewidth]{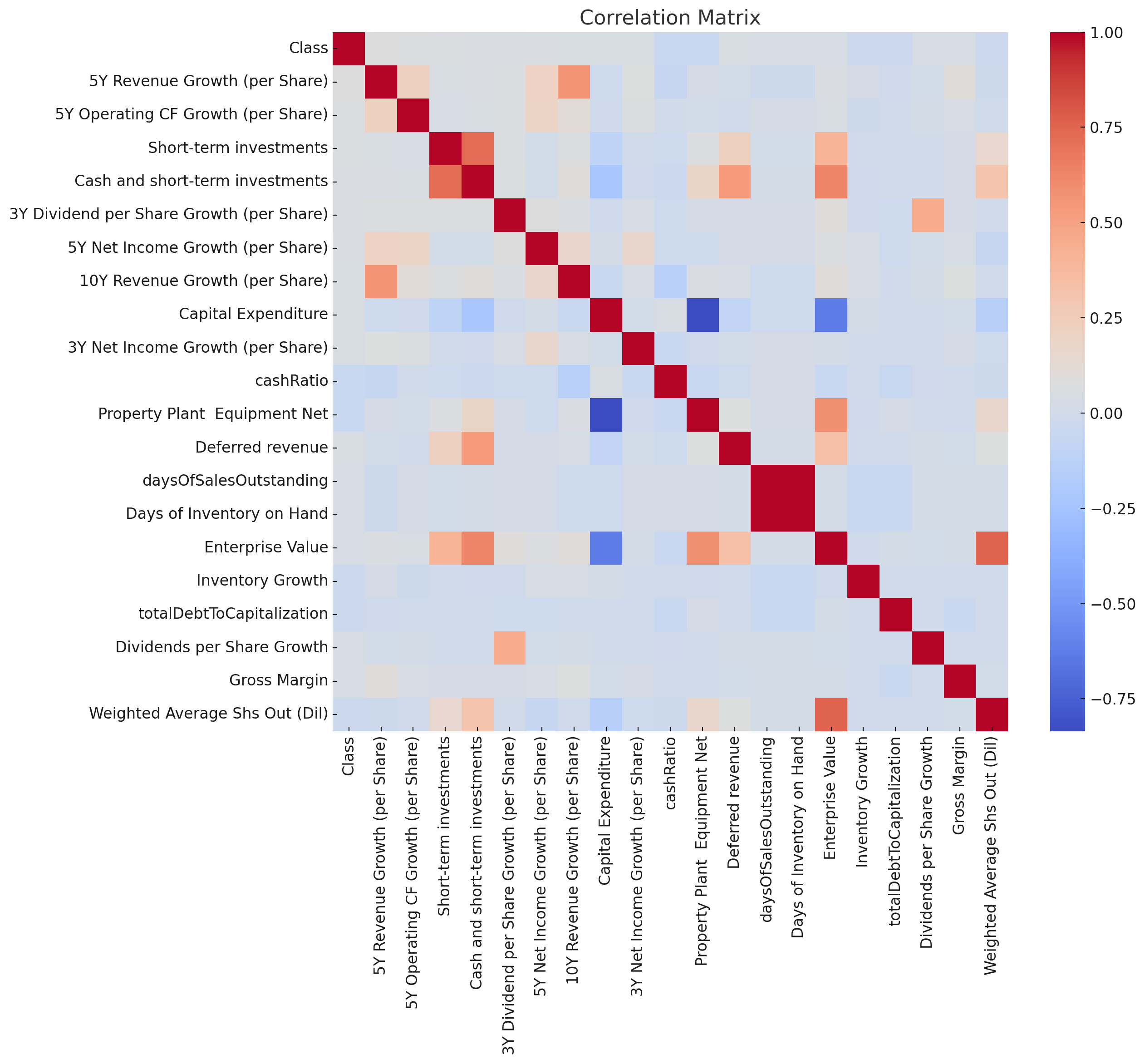}
    \caption{Correlation matrix of selected dataset (year 2014)}
    \label{fig:fig1}
\end{figure}

\subsection{Dataset and experimental setup}

To evaluate the effectiveness of \puck, we selected a diverse set of datasets related to stock market activities spanning the years 2014 to 2018\footnote{\url{https://www.kaggle.com/datasets/cnic92/200-financial-indicators-of-us-stocks-20142018}}. These datasets were chosen due to their highly dynamic nature, real-world complexity, and susceptibility to various types of data errors, such as missing values, noise, outliers, and label misclassifications. 

The dataset collection consists of five different datasets, each containing over 200 financial indicators commonly found in publicly traded company reports. These indicators encompass key financial metrics such as revenue, profit margins, asset values, stock price fluctuations, and trading volumes, offering a comprehensive view of financial market trends. Each dataset includes information on more than 4,000 publicly traded US stocks per year, ensuring a broad representation of market dynamics. However, these datasets are not free from imperfections. Certain financial indicators contain missing values, which can result from incomplete reporting or data collection issues. Additionally, outliers are present, representing extreme values that are likely caused by data entry errors, mistypings, or unusual market behaviors. These underlying issues make the datasets particularly suitable for testing Puck’s ability to simulate realistic data contamination scenarios.

The dataset also includes key financial performance indicators relevant to stock trading decisions. Among these, the second-to-last column, PRICE VAR [\%], represents the percentage price variation of each stock over the course of a given year. This metric is crucial for evaluating a stock’s annual performance. The last column, class, provides a binary classification for each stock based on its price variation. If the PRICE VAR [\%] value is positive, the stock is assigned CLASS = 1, indicating that it has increased in value. From a trading perspective, this signifies that an idealized trader would have benefited from buying the stock at the beginning of the year and selling it at the end for a profit. Conversely, if the PRICE VAR [\%] value is negative, the stock is labeled CLASS = 0, meaning its value declined over the year. In this case, a rational trader would avoid purchasing the stock to prevent capital loss. 
This dataset presents a highly realistic financial scenario, where market fluctuations introduce inconsistencies and imperfections in the data. Stock price variations depend on numerous external factors such as market sentiment, economic conditions, corporate performance, and investor behavior, all of which contribute to data irregularities. Moreover, the presence of missing values, outliers, and misclassified stock performance further enhances the dataset’s suitability for testing \puck’s ability to introduce and manage controlled data contamination.

\begin{figure}[h]
    \centering
    \includegraphics[width=0.8\linewidth]{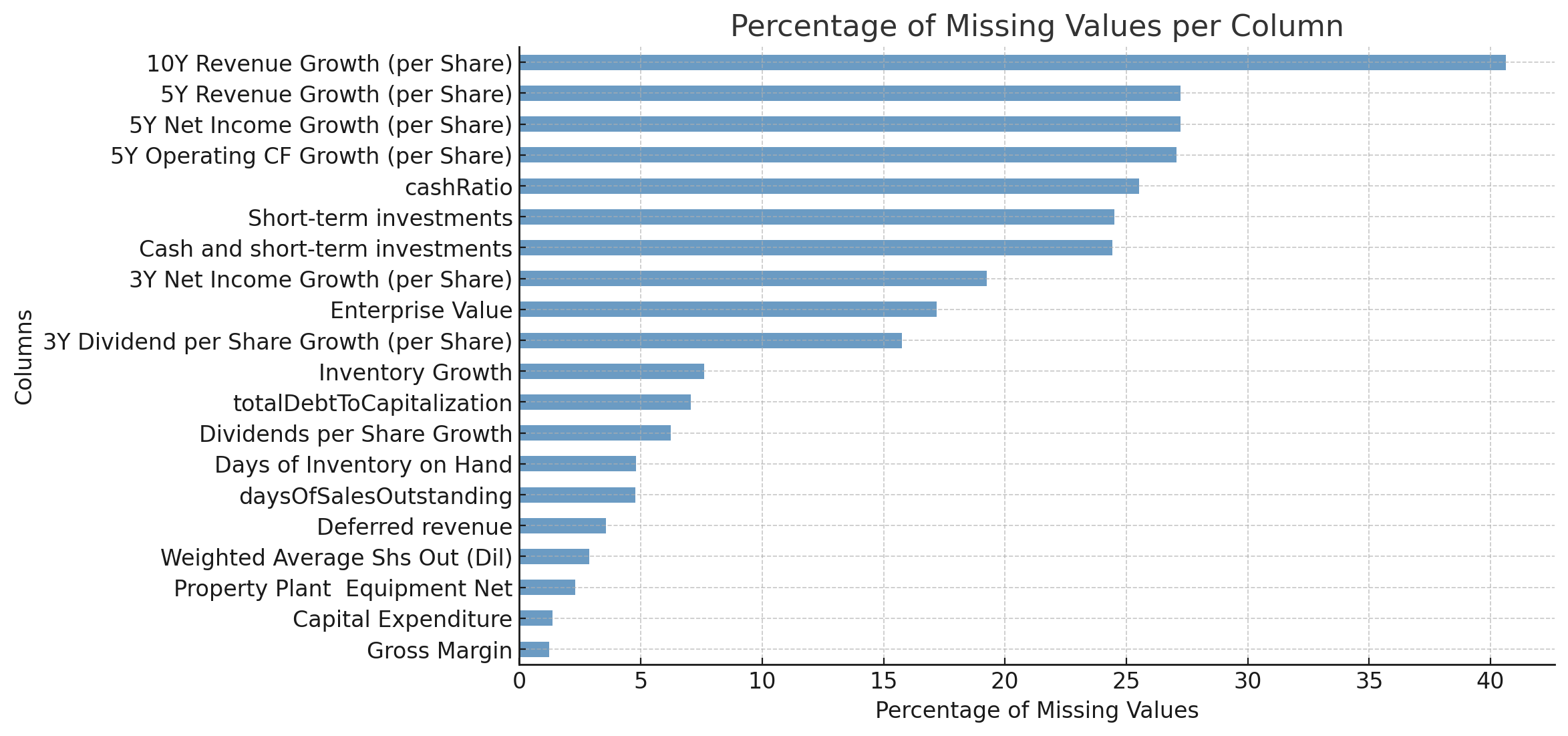}
    \caption{Missing date of the original dataset}
    \label{fig:mising}
\end{figure}

For this experiment, we considered only the top 20 features with the highest correlation with the binary target variable. The correlation matrix of the new dataset is shown in Figure \ref{fig:fig1} highlighting the relationships between financial indicators and stock performance classification. Additionally, the dataset contains a significant number of missing values, as shown in Figure \ref{fig:mising}, which poses challenges for machine learning models and serves as a natural test case for assessing the impact of synthetic data contamination.

To generate the synthetic dataset, we used the GreaT algorithm \cite{DBLP:conf/iclr/BorisovSLPK23} through the publicly available be\_great Python library\footnote{\url{https://github.com/kathrinse/be_great}}. We employed the distilgpt2 large language model with a batch size of 32 for 50 epochs, ensuring a robust generative process capable of mimicking real-world financial data patterns.

To evaluate the effects of controlled data contamination, we created 61 synthetic datasets, categorized as follows:  
\begin{itemize}  
    \item One dataset with only mislabeled errors (error rate fixed at 30\%).  
    \item One dataset for each attribute containing only missing data (error rate fixed at 30\%).  
    \item One dataset for each attribute containing only noisy data (error rate fixed at 30\%).  
    \item One dataset for each attribute containing only outlier data (using the $3\sigma$ model with an error rate fixed at 30\%).  
\end{itemize}

To ensure a comprehensive evaluation, we utilized 10 machine learning models from diverse algorithmic categories, including tree-based models, linear classifiers, probabilistic models, neural networks, and nearest neighbors approaches. The specific models and their corresponding Python implementations are detailed in Table \ref{tab:tab1}. To streamline the model selection, training, and evaluation process, we leveraged PyCaret\footnote{\url{https://pycaret.org/}}, an open-source, low-code Python library that automates machine learning workflows. PyCaret simplifies the experimental pipeline by handling data preprocessing, hyperparameter tuning, and model validation, allowing for efficient large-scale model comparison.

\begin{table}
    \centering
    \begin{tabular}{|l|l|l|}
         \hline
         Label & model Name & Python reference\\ 
         \hline
         SVM &  SVM - Linear Kernel    &  sklearn.linear\_model.\_stochastic\_gradient.SGDC\dots\\ 
         \hline
         ET & Extra Trees Classifier &                sklearn.ensemble.\_forest.ExtraTreesClassifier\\ 
         \hline
         RF & Random Forest Classifier               & sklearn.ensemble.\_forest.RandomForestClassifier \\ 
         \hline
         KN &  K Neighbors Classifier  & sklearn.neighbors.\_classification.KNeighbors\dots\\ 
         \hline
         LDA & Linear Discriminant Analysis     & sklearn.discriminant\_analysis.LinearDiscriminant\dots\\ 
         \hline
         MLP & Multilayer Perception Classifier & sklearn.neural\_network.\_multilayer\_perceptron.\dots\\ 
         \hline
         LR  & Logistic Regression & sklearn.linear\_model.\_logistic.LogisticRegression \\ 
         \hline
         NB &  Naive Bayes & sklearn.naive\_bayes.GaussianNB\\ 
         \hline
         DT &  Decision Tree Classifier &  
 sklearn.tree.\_classes.DecisionTreeClassifier\\ 
         \hline
         QDA & Quadratic Discriminant Analysis  &sklearn.discriminant\_analysis.QuadraticDiscri\dots \\ 
         \hline
    \end{tabular}

    \captionsetup{justification=centerlast, singlelinecheck=true}
    \caption{Machine learning algorithms used in the experimentation.}
    \label{tab:tab1}
\end{table}

\subsection{Results}
By comparing the performance of the 610 developed models (61 contaminated datasets for 10 algorithms), we obtained several key insights into the impact of synthetic data contamination on machine learning accuracy, showing that the use of the \puck library significantly enhances the performance of \ml algorithms compared to using only the synthetic dataset.  

Firstly, we address the question: \textit{Which machine learning algorithm benefits the most from the use of the \puck library?}

To answer this question we analyzed the percentage of models trained on contaminated datasets that achieved a higher accuracy than their counterparts trained on purely synthetic data. The results, summarized in Table \ref{tab:tab4}, reveal key insights into how different machine learning models respond to various types of data contamination. For \textit{noisy data} the Support Vector Machine (SVM) model consistently showed the highest improvement when trained on noise-contaminated datasets across all three years (2014, 2015, and 2016). This suggests that SVM models benefit from controlled noise introduction, possibly due to their inherent ability to find optimal decision boundaries even in the presence of variations in the feature space. Noise may serve as a form of regularization, preventing overfitting and improving model robustness. Regarding \textit{missing values}, Extra Trees (ET) classifier exhibited the most significant improvement when trained on datasets containing missing values. Across all years, ET was the best-performing model for handling missing data, indicating that ensemble tree-based methods are particularly resilient to incomplete datasets. This aligns with previous findings in the literature, as decision trees and ensemble models are capable of handling missing data by splitting based on available features without requiring imputation techniques. Similarly to missing values, ET consistently outperformed other models when trained on datasets containing \textit{outliers}. Interestingly, in 2015, a linear model (ET) also showed competitive performance, suggesting that some models can effectively learn from extreme values when trained on outlier-containing data. This further confirms the robustness of tree-based models in handling anomalies, as they can partition the feature space in a way that minimizes the impact of extreme values. Unlike other types of contamination, no single model stood out as the best performer in handling \textit{mislabeled data}. The results indicate that different models responded differently to label errors, with Linear Discriminant Analysis (LDA) performing well in 2014 dataset, while Random Forest (RF) and Extra Trees (ET) emerged as the strongest models for the 2015 and 2016 datasets, respectively. This suggests that the impact of label misclassification is highly model-dependent, and additional correction mechanisms such as semi-supervised learning or active learning strategies may be needed to mitigate its effects effectively.

\begin{table} [h]
    \centering
    \begin{tabular}{|l|l|l|l|l|l|l|}
    \hline
            &   \multicolumn{2}{|c|}{2014} & \multicolumn{2}{|c|}{2015} & \multicolumn{2}{|c|}{2016}\\ \hline
         Error Type & Model Type & Model& Model Type & Model& Model Type & Model\\
         \hline
         Noisy& Linear & SVM & Linear & SVM & Linear & SVM \\ \hline
         Missing & Tree & ET &Tree & ET & Tree & ET \\ \hline
         Outlier&Tree & ET & Linear & ET & Tree & ET \\ \hline
         Labels & Linear  & LDA & Tree & RF&Tree &ET\\ \hline
       
    \end{tabular}
    \caption{Errors and Model}
    \label{tab:tab4}
\end{table}

A second question we address is: \textit{Which type of error, when introduced into the synthetic dataset, improves or maintains accuracy across all classifiers and throughout all three years of evaluation?}
To answer this question we calculated the percentage of models, trained with contained dataset, with an accuracy that is higher  (best) or higher or equal (best or equal) to the same classifier trained with the synthetic dataset. The results are summarized in Table \ref{tab:tab3}. Across datasets of all years, models trained on noisy data consistently outperformed or matched those trained on purely synthetic data, with 100\% for the 2014 dataset maintaining accuracy and 90\% improving. Although the effect weakened slightly for 2015 and 2016, 70\% of models still maintained or improved performance, confirming that controlled noise acts as a form of regularization, preventing overfitting and enhancing generalization. Missing values also had a positive effect, with 82\% of models in 2014 performing at least as well as those trained on synthetic data, though only 44\% showed direct improvements. Similar trends were observed in 2015 and 2016, where 75\% maintained accuracy, but only 27–35\% improved. This suggests that while some models can adapt to missing data, others require feature engineering or imputation techniques to perform optimally. The impact of outliers was moderate, with 70–79\% of models across all years maintaining or improving accuracy, while 27–39\% experienced direct gains. Though not as influential as noise, outliers were better handled by tree-based models, reinforcing their resilience to high-variance data. Label misclassification had the most unpredictable effect. While 90\% of models maintained or improved accuracy, only 10–20\% saw actual performance gains, indicating that some models can tolerate label noise, while others are significantly affected. The inconsistency suggests that additional correction mechanisms, such as semi-supervised learning or label adjustment strategies, may be necessary to mitigate its impact.

\begin{table} [h]
    \centering
    \begin{tabular}{|l|p{1cm}|l|p{1cm}|l|p{1cm}|l|}
    \hline
            &  \multicolumn{2}{|c|}{2014} & \multicolumn{2}{|c|}{2015} & \multicolumn{2}{|c|}{2016}\\ \hline
         Error Type &	Best or  equal &	Best&	Best or   equal &	Best&	Best or equal &	Best  \\
         \hline
         Labels &  90& 10&\textbf{90}&10&\textbf{90}&20\\ \hline
         Outlier & 79 & 39&76&29&70&27\\
 \hline         Noisy &  \textbf{100}&\textbf{ 90}&70&\textbf{40}&70&\textbf{50} \\ \hline
         Missing&  82 & 44&75&27&75&35\\ \hline
    \end{tabular}
    \caption{Percentage of accuracy improvement wrt. errors }
    \label{tab:tab3}
\end{table}


Table \ref{tab:tab2} presents the percentage of models that exhibited improved accuracy (Best) when trained on \puck contaminated datasets compared to their counterparts trained on purely synthetic data. Additionally, it shows the percentage of models that either improved or maintained accuracy (Best or Equal) relative to the synthetic dataset baseline.

\begin{table}[h]
    \centering
    \begin{tabular}{|l|l|p{1cm}|l|p{1cm}|l|p{1cm}|l|}
    \hline
            & & \multicolumn{2}{|c|}{2014} & \multicolumn{2}{|c|}{2015} & \multicolumn{2}{|c|}{2016}\\ \hline
         Type &	Model &	Best or  equal &	Best&	Best or   equal &	Best&	Best or equal &	Best  \\
         \hline
         \multirow{3}{*}{Tree} &ET & \textbf{96.7} & 95.1 & \textbf{98.36}& 85.25&\textbf{91.80}&55.74\\
          & RF& 95.1& 91.8& 88.52&9.84&44.26&8.20\\         
          & DT &85.2& 70.5& 44.26 &24.59&72.13&54.10\\
         \hline
         \multirow{2}{*}{Linear} & LDA & 93.4 &83.6& \textbf{95.08}&9.84&86.89&11.48\\
          &	SVM & \textbf{100} &49.2&93.44&42.62&\textbf{96.72}&52.46\\
          & LR & 91,8 & 45.9&42.62& 6.56&91.80&47.54\\
         \hline
         Neural &MLP & \textbf{93.4} & 83.6&\textbf{85.25}&42.62&\textbf{95.08}&52.46\\
         \hline
         \multirow{2}{*}{Probabilistic} &QDA& 65.6& 63.9&\textbf{73.77}&65.57&3.28&3.28\\
         & NB & \textbf{86.9}&1.6&63.93&19.67&\textbf{90.1}6&55.74\\
         \hline
         Nieghbors &KN& \textbf{95.1} & 47.5&\textbf{50.82}&3.28&\textbf{45.90}&24.59\\
         \hline
        
    \end{tabular}
    \caption{Percentage of improvement of accuracy by model and type }
    \label{tab:tab2}
\end{table}

As shown in Table \ref{tab:tab2}, Puck enhances the performance of Extra Trees and Random Forests in over 90\% of experiments, demonstrating its effectiveness in improving tree-based models. Furthermore, \puck consistently improves or maintains the accuracy of models trained on the original synthetic dataset, ensuring that data contamination does not degrade performance. In general, the results confirm that \puck positively impacts or preserves accuracy across all major \ml algorithm categories, including tree-based, linear, neural, and neighbor-based models.


\section{Conclusions}
\label{ref:conclusion}
Training machine learning models with real-world data is often difficult due to privacy and confidentiality restrictions, limiting the ability to develop classifiers using actual datasets. To address this, various synthetic data generation algorithms have been developed over the years, aiming to replicate the statistical properties of real data. However, a critical limitation of synthetic datasets is their lack of real-world imperfections. Unlike actual datasets, synthetic data is often too clean, as it does not contain common data issues such as missing values, outliers, and label misclassification, which are inherent in real-world applications.
In this paper we introduced \puck a library designed to systematically introduce a controlled percentage of errors into datasets, either at the feature level or across the entire dataset. Experimental evaluations on three real-world datasets confirm the effectiveness of this approach. Notably, linear models such as SVMs and tree-based models demonstrate improved accuracy when trained on controlled-error datasets, suggesting that strategic contamination can enhance model robustness.

There are several directions for future research. One key focus is the development of an enhanced version of the \puck library that supports a more generalized strategy for error insertion, allowing for greater flexibility in simulating real-world data imperfections. Additionally, we are exploring methods to modify dataset schemas rather than only altering data values. A common challenge arises when the schema of a dataset used in a machine learning model differs from that of the training dataset. For instance, a feature such as total living area in the training data may later be divided into interior area and exterior area in a real-world application, leading to a structural mismatch that could affect model performance. Addressing such discrepancies will be crucial for ensuring model adaptability across different data environments.

Furthermore, additional experiments are needed to gain deeper insights into the relationship between error types, model architectures, and feature selection. A key objective is to help users identify the minimum set of essential features required to improve classifier accuracy, even when working with imperfect datasets. By refining these aspects, we aim to make \puck a more comprehensive tool for evaluating machine learning models under realistic data conditions.

\end{document}